\documentclass[letterpaper]{article} % DO NOT CHANGE THIS
\usepackage{aaai2027}  % DO NOT CHANGE THIS
\nocopyright
\usepackage[hyphens]{url}  % DO NOT CHANGE THIS
\usepackage{graphicx} % DO NOT CHANGE THIS
\usepackage{natbib}  % DO NOT CHANGE THIS AND DO NOT ADD ANY OPTIONS TO IT
\usepackage{caption} % DO NOT CHANGE THIS AND DO NOT ADD ANY OPTIONS TO IT
\usepackage{algorithm}
\usepackage{algorithmic}

\usepackage{newfloat}
\usepackage{listings}
\DeclareCaptionStyle{ruled}{labelfont=normalfont,labelsep=colon,strut=off} % DO NOT CHANGE THIS
\floatstyle{ruled}
\newfloat{listing}{tb}{lst}{}
\floatname{listing}{Listing}

\usepackage{booktabs}

\usepackage{amssymb}
\usepackage{amsmath}
\usepackage{multirow}
\usepackage[capitalize]{cleveref}

\title{Estimating Uncertainty in Galaxy Morphology Classification}
\author{
    Kai Cheng\textsuperscript{\rm 1},
    Ruoqi Wang\textsuperscript{\rm 2},
    Qiong Luo\textsuperscript{\rm 1,2}
}
\affiliations{
    \textsuperscript{\rm 1}The Hong Kong University of Science and Technology, Hong Kong, China\\
    \textsuperscript{\rm 2}The Hong Kong University of Science and Technology (Guangzhou), Guangzhou, China\\
}

\begin{document}

\maketitle

\begin{abstract}
Astronomers classify galaxy morphology to investigate cosmic evolution. While deep foundation models are increasingly utilized in Galaxy Morphology Classification (GMC), little work has been done on evaluating the uncertainty of GMC results. Uncertainty evaluation is important because astronomical data are inherently noisy due to instrumental and environmental limitations. Also, the continuous evolution of galaxies creates intrinsic morphological ambiguity. However, current foundation models operate as deterministic point estimators, failing to quantify the uncertainty. To overcome this limitation, we propose UEGMC, a post-hoc framework of Uncertainty Estimation for Galaxy Morphology Classification. It categorizes uncertainty in GMC into distinct types by model parameters, astronomical data, reference standards, or intrinsic physical ambiguities, thereby facilitating better classification. Our framework can directly predict uncertainties from representations extracted from the frozen backbones of foundation models, without computationally expensive sampling, therefore enabling fine-grained uncertainty evaluations. Our experimental results demonstrate that UEGMC provides competitive uncertainty quantification performance compared with previous methods.
\end{abstract}


\section{Introduction}
\begin{figure}[t]
    \centering
    \includegraphics[width=\columnwidth]{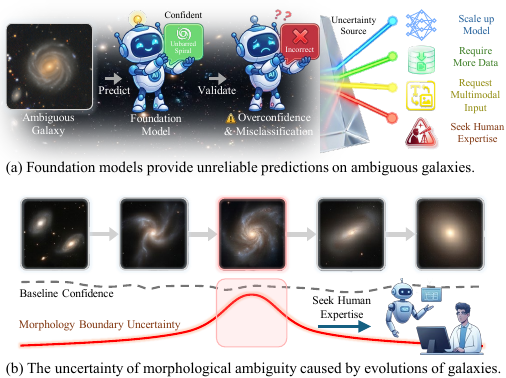}
    \caption{Unaddressed uncertainty in foundation models and the critical role of cognitive identification for galaxy morphology classification.}
    \label{fig:fig1}
\end{figure}

Classifying the morphology of galaxies is important for astronomers \cite{baker2025core, wang2025giant}, providing critical insights into cosmics \cite{angeloudi2024constraints}. As astronomical surveys continuously increase, deep foundation models \cite{lastufka2024vision} have been rapidly applied to Galaxy Morphology Classification (GMC), taking advantage of their powerful representation learning capabilities across diverse data \cite{abdullah2026tap}. Despite predictive success, current GMC results are missing certainty information. As shown in Figure ~\ref{fig:fig1} (a), recent foundation models for GMC typically lack a quantitative measure of their confidence \cite{abdar2021review}. Challenged by noise arising from instrumental limitations and intricate physical mechanisms \cite{wang2025galaxalign}, these models are prone to misclassifications with overconfidence \cite{dominguez2019transfer, kim2025generalized}. The inability to trace the complex sources of uncertainty restricts these models. Therefore, Uncertainty Quantification (UQ) \cite{bethell2024robust} is vital to establish scientific credibility bounds for GMC results. In this paper, we propose a framework for UQ in GMC.

\begin{figure}[t]
    \centering
    \includegraphics[width=\columnwidth]{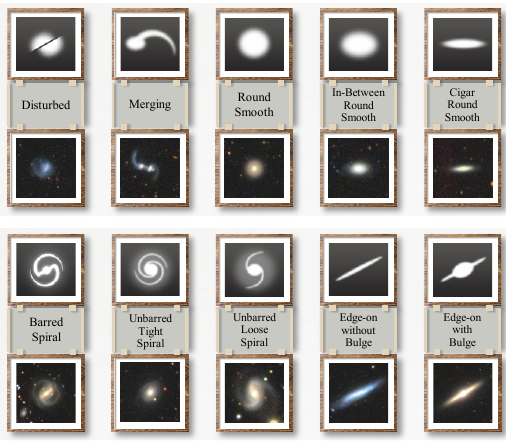}
    \caption{Illustration of galaxy morphology description and schematic symbols corresponding to the examples of galaxy images. In each morphology type, the upper and lower panels show the schematic symbols and corresponding examples of galaxy images, respectively. Images and textual labels are from the Galaxy10 DECaLS dataset \cite{Galaxy10}, whereas schematic symbols are from the Galaxy Zoo 2 decision tree \cite{willett2013galaxy,dieleman2015rotation}.}
    \label{fig:fig2}
\end{figure}

Quantifying the confidence in GMC presents two unique challenges. The main challenge is that the underlying multiple sources of uncertainty are complex \cite{kendall2017uncertainties}, rendering a single homogeneous uncertainty evaluator inadequate. First, galaxy data are inherently imperfect due to instrumental limitations and environmental factors during collection \cite{lastufka2024vision, slijepcevic2024radio}. Second, many approaches \cite{mishra2024paperclip, wang2025galaxalign} emulate citizen scientists to achieve multimodal GMC by incorporating supplementary reference standards, as shown in Figure~\ref{fig:fig2}. Such multimodal integration introduces deviations in understanding standards from multiple angles \cite{chen2025acknowledging}. The symbols are highly idealized, creating a domain gap with physical images. Third, galaxies frequently reside in transitional states due to continuous morphological evolutions \cite{walmsley2022practical, baker2025core}, as shown in Figure~\ref{fig:fig1} (b), causing an intrinsic physical boundary blur.

The second challenge lies in the limitations of calculating the predictive uncertainty. On the one hand, although sampling-based methods, such as Bayesian Neural Networks (BNNs) \cite{goan2020bayesian}, can empirically estimate the uncertainty, they are highly cost-intensive and structurally incompatible with post-hoc evaluations. On the other hand, previous prediction-based methods are limited in distinguishing uncertainties originating from diverse sources for GMC \cite{chun2026evidential, mohammed2026gem}.

To address these challenges, we introduce Uncertainty Estimation for Galaxy Morphology Classification (UEGMC), a post-hoc framework that empowers foundation models with a fine-grained uncertainty quantification capability by categorizing the predictive uncertainty into distinct types, as shown in Figure~\ref{fig:fig1} (a). UEGMC first establishes a baseline, called UEGMC-B, by performing Bayesian approximations across four uncertainty types in our categorization. Then, we propose UEGMC-P, encompassing a suite of strategies designed to achieve sampling-free and lightweight uncertainty predictions and alleviate computational costs.

We categorize the predictive uncertainty as follows. First, the model epistemic \cite{depeweg2018decomposition} uncertainty reflects the cognitive capacity of the foundation model, which cost-efficiently helps evaluate whether scaling to a larger model will be worthwhile. Second, the data aleatoric \cite{depeweg2018decomposition} uncertainty evaluates the quality of data \cite{kendall2017uncertainties}. Because astronomical data exhibit long-tail distributions and scarcity \cite{hosenie2020imbalance}, the training data cannot fully cover the vast diversity of astrophysical phenomena. While pretraining on extensive datasets is deployed to alleviate this issue \cite{oquab2023dinov2, walmsley2023zoobot}, retrieving this aleatoric component provides a metric to assess whether such computationally expensive pretraining will yield meaningful improvements in classification. Third, the modality aleatoric uncertainty quantifies the deviations introduced when foundation models comprehend supplementary reference standards. Finally, the morphology boundary uncertainty captures the ambiguity caused by intricate morphological evolutions. Thus, our framework empowers the model to recognize its main sources of limitations and low-confidence predictions. Also, it enables the quantification of physical uncertainty that is challenging to capture, allowing physically ambiguous samples to be effectively deferred to astronomers and thus preserving human effort.

Our contributions are as follows:
\begin{itemize}
    \item By explicitly categorizing uncertainty from multiple sources, we propose the UEGMC framework to perform a fine-grained uncertainty quantification tailored to model the complex nature of galaxy morphology.
    \item Our UEGMC not only provides a Bayesian approximation baseline, but also proposes the post-hoc, efficient, and sampling-free UEGMC-P to alleviate the issue of computational limitations and costs.
    \item Extensive experimental results demonstrate the effectiveness and robustness of our proposed UEGMC in accurately capturing and quantifying distinct sources of uncertainty in GMC.
\end{itemize}
\section{Background and Related Work}

\subsection{Foundation Models in Astrophysics}
Modern astrophysical surveys produce a large amount of data. Unlike natural images, they exhibit systematic noise and artifacts alongside a highly dynamic range \cite{wang2025galaxalign}, prompting continuous exploration into the application of foundation models in this domain \cite{lastufka2024vision}. Moreover, existing models designed for astronomical tasks, including Zoobot \cite{walmsley2023zoobot} and CLIP-based methods \cite{mishra2024paperclip,wang2025galaxalign}, focus on scaling datasets or applying multi-modal pretraining to enhance the performance. However, previous work does not address the uncertainty of GMC. In contrast, we explicitly identify and model the inherent uncertainties from multiple sources to better facilitate classification.

\subsection{Uncertainty Quantification}
Quantifying uncertainty in deep neural networks traditionally relied on computationally expensive methods such as BNNs \cite{goan2020bayesian} or Deep Ensembles \cite{lakshminarayanan2017simple}. Recently, some post-hoc Artificial Intelligence (AI) methods have been proposed. Deng \textit{et al.} \cite{deng2023uncertainty} utilized the Fisher information matrix to dynamically reweight loss terms for improved uncertainty estimation. Chen \textit{et al.} \cite{chen2024r} treated the prior weight as an adjustable hyperparameter and directly optimized the expected Dirichlet distribution without variance-minimizing regularization. Mohammed \textit{et al.} \cite{mohammed2026gem} proposed gated evidential mixtures, which integrated a distance-informed gating mechanism with a Fisher-regularized mixture of evidential heads. Chun \textit{et al.} \cite{chun2026evidential} proposed to convert standard pretrained models into evidential models by interpreting the transformed output as parameters of a Dirichlet distribution for uncertainty estimation. These methods did not categorize various types of uncertainty critical to GMC. In contrast, our UEGMC offers a physics-aware uncertainty categorization, and the computational overhead is low.
\section{Methodology}
\label{sec:method}

\begin{figure*}[ht]
    \centering
    \includegraphics[width=\linewidth]{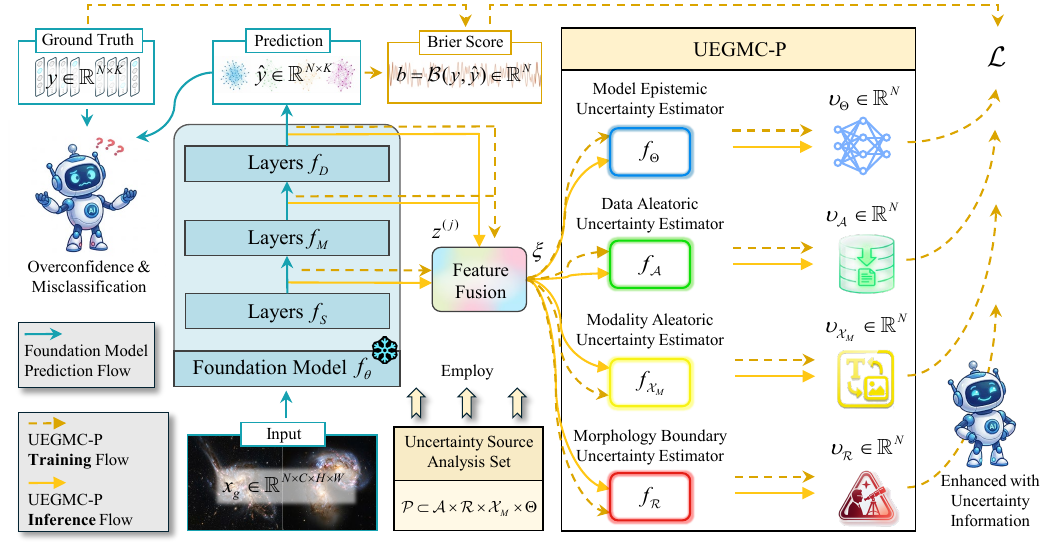}
    \caption{An illustration of our proposed UEGMC-P framework. With the foundation model kept frozen, UEGMC-P undergoes independent training in a post-hoc manner to provide categorized uncertainty predictions across four types. This mechanism evaluates the overconfidence inherent in the direct output of the original model and its sources of uncertainty.}
    \label{fig:fig3}
\end{figure*}

In this section, we present UEGMC, a framework we specifically design to categorize and predict critical uncertainties for GMC. First, we perform a fine-grained categorization of the uncertainty, facilitating corresponding quantification strategies for uncertainties from distinct sources. Subsequently, we leverage established foundation models to conduct Bayesian approximations, thereby providing a baseline and revealing their limitations. Finally, we present UEGMC-P, our post-hoc training approach to sampling-free uncertainty prediction for GMC.

\subsection{Uncertainty Categorization for GMC}
We denote the entire parameter set of the model as $\theta$ and the training data as $\mathcal{D}$. Given an input $x$, the comprehensive Bayesian predictive distribution \cite{gal2015bayesian} for the target $y$ is formulated as follows:
\begin{equation} \label{eq:integral1}
    p(y | x, \mathcal{D}) = \int_{\theta} p(y | x, \theta) p(\theta | \mathcal{D}) d\theta.
\end{equation}
This integral provides a basis for quantifying the confidence that the prediction of the model is correct. Typically, the term $p(y | x, \theta)$ is categorized into the \textit{Aleatoric} uncertainty, while $p(\theta | \mathcal{D})$ is the \textit{Epistemic} uncertainty \cite{depeweg2018decomposition}. However, it also presents significant computational challenges because acquiring the posterior distribution of the model parameters $p(\theta | \mathcal{D})$ is typically intractable \cite{liu2023simple}, and integrating over the vast parameter space requires excessive computational resources.

To facilitate a finer-grained categorization to match various sources in GMC, we further expand Equation \eqref{eq:integral1}, the predictive distribution, into multiple integrals involving distinct probabilistic components as follows:
\begin{equation} \label{eq:integral2}
    % \resizebox{.43\textwidth}{!}{$
    \begin{aligned}
    p(y | x, & \mathcal{D}) = \int_{m} \int_{x_M} \int_{\theta} \, \underbrace{p(y | m, \theta, x)}_{\text{Data Aleatoric}} \\
    & \underbrace{p(m | \theta, x)}_{\substack{\text{Morphology} \\ \text{Boundary}}} \underbrace{p(x_M | x_g)}_{\substack{\text{Modality} \\ \text{Aleatoric}}} \underbrace{p(\theta | \mathcal{D})}_{\substack{\text{Model} \\ \text{Epistemic}}} \, d\theta dx_M dm
    \end{aligned} \, ,
    % $},
\end{equation}
where $x_g$ is the galaxy image, $x_M$ is the supplementary multimodal data, $x = \left\{ x_g, x_M \right\}$ is the input, and $m$ is the latent variable for modeling the physical boundary. These four components systemically categorize the predictive uncertainty into four types. 

Specifically, the posterior distribution $p(\theta | \mathcal{D})$ represents the model epistemic uncertainty. Since reference standards, including schematic symbols and textual descriptions, are exploited in several GMC approaches \cite{mishra2024paperclip,wang2025galaxalign} to characterize the intrinsic cognitive process of citizen scientists, it is essential to account for multimodal implications. Thus, we divide the \textit{Aleatoric} uncertainty by the added modalities. Formally, $x$ is decomposed into $x_M$ and the original galaxy image $x_g$, which is denoted as $x = \left\{ x_g, x_M \right\}$. This way, the relationship $p(x_M | x_g)$ between $x_g$ and $x_M$ represents the modality aleatoric uncertainty. And the model outputs the prediction probability $p(y | m, \theta, x)$ given the input data $x$, reflecting the data aleatoric uncertainty. Furthermore, we incorporate a latent variable $m$ to model the physical boundary, morphological ambiguity caused by the evolution of the galaxy. In other words, by modeling $p(m | \theta, x)$ in our UEGMC, we can quantitatively assess the morphology boundary uncertainty. 

\subsection{UEGMC-B: Bayesian Approximation}
The predictive uncertainty in Equation \eqref{eq:integral2} is intractable \cite{liu2023simple}. To tackle this issue, we first consider sampling-based methods \cite{liu2023simple, wang2025review} to approximate the uncertainty. We utilize the Shannon entropy of the predictive probability distribution to realize the Bayesian UQ. For a classification task with $N$ categories, the entropy of a prediction is formulated as:
\begin{equation}
    H(y) = - \sum_{i=1}^{N} y_i \log_2(y_i). 
\end{equation}
We define a set $\mathcal{Y}=\left\{ y^i \right\}^K_{i=1}$ , where $K$ is the product of the cardinalities of four distinct sets $\Theta$, $\mathcal{A}$, $\mathcal{X}_M$, and $\mathcal{R}$. Specifically, $\theta \in \Theta$ represents models of various sizes. $a \in \mathcal{A}$ represents pretraining on different external datasets, encompassing both general and domain-specific data enhancements. $x_M \in \mathcal{X}_M$ denotes different configurations for integrating supplementary modalities $\left\{\emptyset, x_{txt}, x_{sym}, \left\{x_{txt}, x_{sym} \right\}\right\} $, where $x_{sym}$ and $x_{txt}$ denote schematic symbols and textual descriptions, respectively. Finally, $r \in \mathcal{R}$ denotes a series of galaxy evolution phenomena since galaxies may exist in transitional evolving states and cause ambiguity. Therefore, we formulate a physics-consistent strategy to reveal morphological ambiguity by performing perturbations on the labels of samples based on the phenomena of galaxy evolution. For instance, there is a phenomenon "Unbarred Loose Spiral $\rightarrow$ Unbarred Tight Spiral", which is driven by central mass accumulation and enhances the shear rate of differential rotation. This phenomenon leads to a reduction in the spiral pitch angle, resulting in tightly wound spiral arms. Thus, a sample originally labeled as "Unbarred Loose Spiral" can be subjected to a perturbation that shifts the label to "Unbarred Tight Spiral".

To evaluate the model epistemic uncertainty caused by insufficient learning capabilities, we fix the data and labels to their unperturbed states and allow only the model parameters $\theta$ to vary within the set $\Theta$. The corresponding entropy is approximated by the expected prediction weighted by the probability $p(\theta | \mathcal{D})$ as follows: 
\begin{equation}
    \resizebox{.42\textwidth}{!}{$
    H_{\Theta}\left(x_g,r,x_M, \Theta, \mathcal{Y}\right) \approx H\left( \sum\limits_{\theta \in \Theta} p\left( \theta | \mathcal{D} \right) \cdot y_{x_g,r,x_M,\theta} \right)
    $},
\end{equation}
where $p(\theta | \mathcal{D}) = \frac{1}{|\Theta|}$.

Similarly, the data aleatoric uncertainty is estimated as follows:
\begin{equation}
    \resizebox{.42\textwidth}{!}{$
    H_{\mathcal{A}}\left(x_g,r,x_M, \theta, \mathcal{A}, \mathcal{Y}\right) \approx H\left( \sum\limits_{a \in \mathcal{A}} p\left( y | x, \theta, m \right) \cdot y_{a(x_g),r,x_M,\theta} \right)
    $},
\end{equation}
where $p(y | x, \theta, m) = \frac{1}{|\mathcal{A}|}$.

The modality aleatoric uncertainty is assessed by varying the multimodal configurations within $\mathcal{X}_M$ as follows:
\begin{equation}
    \resizebox{.42\textwidth}{!}{$
    H_{\mathcal{X}_M}\left(x_g,r,\theta,\mathcal{X}_M, \mathcal{Y}\right) \approx H\left( \sum\limits_{x_M \in \mathcal{X}_M} p\left( x_M | x_g \right) \cdot y_{x_g,r,x_M,\theta} \right)
    $},
\end{equation}
where $p(x_M | x_g) = \frac{1}{|\mathcal{X}_M|}$.

Finally, the morphology boundary uncertainty is estimated as follows:
\begin{equation}
    \resizebox{.42\textwidth}{!}{$
    H_{\mathcal{X}_M}\left(x_g,x_M, \theta, \mathcal{R}, \mathcal{Y}\right) \approx H\left( \sum\limits_{r \in \mathcal{R}} p\left( m | x, \theta \right) \cdot y_{x_g,r,x_M,\theta} \right)
    $},
\end{equation}
where $p(m | x, \theta) = \frac{1}{|\mathcal{R}|}$. Directly executing this Bayesian entropy approximation is computationally expensive and demands ground truth information when evaluating the morphology boundary uncertainty. Consequently, we deploy this Bayesian methodology within our framework as a comparative baseline only, designated as UEGMC-B.

\subsection{UEGMC-P: Uncertainty Prediction}
Through the derivation of the Bayesian approach, the extensive multiple integral problem is transformed into a series of discrete expected values. However, Monte Carlo variations used to estimate uncertainty are not applicable post-hoc and are highly costly. To address this issue, we train lightweight estimators to efficiently predict uncertainties from the latent space. As shown in Figure ~\ref{fig:fig3}, foundation models only generate a single label prediction $\hat{y}$, lacking any quantitative measure of the model's confidence. In contrast, our UEGMC-P leverages the hierarchical embeddings $z^{(j)}$ to achieve feature fusion $\xi = \text{Concat}(z^{(j)})$, ensuring that UEGMC-P simultaneously yields the nature of the observation and the current predictive inclination. The UEGMC-P estimators employ the lightweight Multi-Layer Perceptron (MLP) architecture tailored to map $\xi$ to the target uncertainty estimations. Also, UEGMC-P is post-hoc and freezes the converged foundation models, ensuring that only the weights of the estimators are updated via backpropagation. Our design not only guarantees that the estimators fit the target uncertainties in a stable representation space, but also keeps the uninterrupted feature learning process required for classification. 

Due to the distinct characteristics of uncertainties, we construct different learning tasks and strategies for their respective estimators. The Brier score serves as the main metric to evaluate the discrepancy between the prediction $\hat y$ and the corresponding ground truth $y$, calculated as:
\begin{equation}
    BS \left( \hat{y}^i, y^i \right) = \sum_{i=1}^{K} (\hat{y}^i - y^i)^2.
\end{equation}
where $K$ is the total number of categories. The Brier score simultaneously measures reliability and uncertainty \cite{murphy1973new}, penalizing predictions that are both overconfident and incorrect. 

We formulate a function $\rho$ to map $\xi$ to the respective Brier scores. First, the model epistemic uncertainty is quantified as the difference in predictions between the estimators trained on the results from tiny and large model architectures, denoted as $\upsilon_{\Theta} = \rho_{T}(\xi) - \rho_{L}(\xi)$. Second, we assess the data aleatoric uncertainty utilizing models fine-tuned after general or domain-specific pretraining. The average Brier scores from these models define the target of $\rho_{\mathcal{A}}(\xi)$. The corresponding data aleatoric uncertainty is measured as $\upsilon_{\mathcal{A}} = \rho_{B}(\xi) - \rho_{\mathcal{A}}(\xi)$, where $\rho_{B} \left(\xi\right)$ represents the estimator trained on the results from the base model. Third, to quantify the modality aleatoric uncertainty, we incorporate configurations of supplementary modalities, $\left\{\emptyset, x_{txt}, x_{sym}, \left\{x_{txt}, x_{sym} \right\}\right\} $, to calculate their expected Brier scores. The results are then averaged to serve as the target of the estimator $\rho_{\mathcal{X}_M}$. The modality aleatoric uncertainty is consequently quantified by $\upsilon_{\mathcal{X}_M} = \rho_{B}(\xi) - \rho_{\mathcal{X}_M}(\xi)$. Finally, we calculate auxiliary Brier Scores using the changed labels of galaxy morphologies into which the sample could potentially evolve. The morphology boundary uncertainty can be quantified as $\upsilon_m = 1 - |\rho_{B}(\xi) - \rho_m(\xi)|$.

\section{Experiments}
\label{sec:exp}

\subsection{Experimental Setup}
All proposed models and experiments are implemented in PyTorch \cite{paszke2019pytorch}. The training and evaluation processes are conducted on a high-performance Linux workstation equipped with an NVIDIA H100 GPU. In our experiments, we evaluate our UEGMC on two widely adopted public galaxy datasets: (1) Galaxy10 DECaLS \cite{Galaxy10} and (2) GalaxyMNIST \cite{GalaxyMNIST}.

\subsubsection{Foundation Models}
To comprehensively evaluate our approach and facilitate comparison with previous UQ methods, we employ a diverse suite of foundation models as backbones, encompassing two mainstream architectures: (1) Vision Transformers (ViT) \cite{DBLP:conf/iclr/DosovitskiyB0WZ21} and (2) ConvNeXt \cite{woo2023convnext}. This suite comprises models of diverse scales, a variety of unimodal and multimodal models, and models pretrained on general and domain-specific datasets.

\begin{table}[h]
    \centering
    \caption{Comparison of the model epistemic uncertainty quantification across datasets with the ViT \cite{DBLP:conf/iclr/DosovitskiyB0WZ21} backbone. This table shows the relative Area Under Curve (AUC) (\%) $\uparrow$ when predicting a variable fraction of the most uncertain samples with a larger model, while predictions of other samples remain unchanged. \textbf{Bold} and \underline{underlined} numbers denote the best and second best results respectively.}
    \resizebox{\columnwidth}{!}{
    \begin{tabular}{l|ccc|ccc}
        \toprule
        \multirow{3}{*}{Method} & \multicolumn{3}{c}{GalaxyMNIST} & \multicolumn{3}{c}{Galaxy10} \\ \cmidrule(lr){2-4} \cmidrule(lr){5-7}
        & AUC & AUC & AUC & AUC & AUC & AUC \\ 
        & (Acc) & (F1 Score) & (mAP) & (Acc) & (F1 Score) & (mAP) \\ 
        \midrule
        DeepEns & 28.659 & 28.210 & 11.851 & 24.345 & 22.721 & 21.863 \\
        \textit{I}-EDL & 26.882 & 27.142 & 11.733 & 23.913 & 22.236 & 21.097 \\
        R-EDL & 27.086 & 25.902 & 11.871 & \underline{25.834} & 23.669 & 20.962 \\
        GEM-FI & 27.120  & 27.101 & 11.654 & 25.213 & 23.283 & 20.347 \\
        ETN & \underline{29.215} & \underline{29.307} & \underline{11.905} & 23.563 & 23.232 & 21.852 \\
        MaxLogit & 27.016 & 26.996 & 11.779 & 23.791 & 24.006 & \underline{22.673} \\ \cmidrule(lr){1-7}
        \textbf{UEGMC-B} & 25.550 & 25.525 & 11.779 & 24.587 & \underline{25.555} & 22.354 \\
        \textbf{UEGMC-P} & \textbf{41.885} & \textbf{41.807} & \textbf{12.281} & \textbf{43.199} & \textbf{41.353} & \textbf{36.633} \\
        \bottomrule
    \end{tabular}
    }
    \label{tab:tab1}
\end{table}

\begin{table}[t]
    \centering
    \caption{Comparison of the data aleatoric uncertainty quantification across datasets with the ViT \cite{DBLP:conf/iclr/DosovitskiyB0WZ21} backbone. This table shows the relative Area Under Curve (AUC) (\%) $\uparrow$ when predicting a variable fraction of the most uncertain samples with more pretrained data, while predictions of other samples remain unchanged. \textbf{Bold} and \underline{underlined} numbers denote the best and second best results respectively.}
    \resizebox{\columnwidth}{!}{
    \begin{tabular}{l|ccc|ccc}
        \toprule
        \multirow{3}{*}{Method} & \multicolumn{3}{c}{GalaxyMNIST} & \multicolumn{3}{c}{Galaxy10} \\ \cmidrule(lr){2-4} \cmidrule(lr){5-7}
        & AUC & AUC & AUC & AUC & AUC & AUC \\ 
        & (Acc) & (F1 Score) & (mAP) & (Acc) & (F1 Score) & (mAP) \\ 
        \midrule
        DeepEns & 34.076 & 34.324 & 36.557 & 35.478 & 41.324 & 44.213 \\
        \textit{I}-EDL & 46.799 & 43.518 & 57.425 & 44.996 & 44.109 & 46.136 \\
        R-EDL & 43.973 & 45.841 & 57.591 & 45.704 & 43.951 & 49.407 \\
        GEM-FI & 45.231 & 45.104 & 56.510 & 43.915 & 42.817 & 48.259 \\
        ETN & 34.846 & 34.850 & 39.714 & 36.613 & 41.975 & 44.610 \\
        MaxLogit & 81.462 & 81.573 & 85.026 & 72.087 & 71.455 & 75.567 \\ \cmidrule(lr){1-7}
        \textbf{UEGMC-B} & \underline{92.615} & \underline{92.598} & \underline{89.063} & \underline{82.496} & \underline{83.060} & \underline{81.647} \\
        \textbf{UEGMC-P} & \textbf{94.538} & \textbf{94.526} & \textbf{91.276} & \textbf{89.643} & \textbf{88.489} & \textbf{89.497} \\
        \bottomrule
    \end{tabular}
    }
    \label{tab:tab2}
\end{table}

\subsubsection{UQ Methods under Comparison}
The comparison UQ methods include Deep Ensembles \cite{lakshminarayanan2017simple}, \textit{I}-EDL\cite{deng2023uncertainty},  R-EDL\cite{chen2024r}, ETN \cite{chun2026evidential}, and GEM-FI \cite{mohammed2026gem}. To evaluate the effectiveness of our proposed categorization strategies, we also compare with using the maximum logit value as an indicator of uncertainty.

\subsubsection{Evaluation Metrics}
We select three standard metrics in classification tasks, including Accuracy (Acc), F1 Score, and mean Average Precision (mAP). While these metrics quantify the deterministic point-estimation performance of a model, we evaluate and rank competing UQ methods based on the Area Under Curve (AUC) derived from these indicators \cite{kendall2017uncertainties, geifmanbias}. Specifically, across the four uncertainty types, we compute the AUC by identifying a variable fraction between 0\% and 100\% of the most uncertain samples and substituting their original estimates with a better one. Depending on the specific type under evaluation, we either correct these uncertain samples using ground-truth labels to simulate human expert intervention or refine their predictions by using a larger model, leveraging extensive pretraining datasets, or incorporating additional data modalities. An effective UQ method systematically assigns higher uncertainty scores to samples most vulnerable to misclassification. Thus, targeted refinement of these samples improves classification performance, increasing AUC and demonstrating its efficacy.

\begin{table}[t]
    \centering
    \caption{Comparison of the modality aleatoric uncertainty quantification across datasets with the ViT \cite{DBLP:conf/iclr/DosovitskiyB0WZ21} backbone. This table shows the relative Area Under Curve (AUC) (\%) $\uparrow$ when predicting a variable fraction of the most uncertain samples with more combinations of modalities, while predictions of other samples remain unchanged. \textbf{Bold} and \underline{underlined} numbers denote the best and second best results respectively.}
    \resizebox{\columnwidth}{!}{
    \begin{tabular}{l|ccc|ccc}
        \toprule
        \multirow{3}{*}{Method} & \multicolumn{3}{c}{GalaxyMNIST} & \multicolumn{3}{c}{Galaxy10} \\ \cmidrule(lr){2-4} \cmidrule(lr){5-7}
        & AUC & AUC & AUC & AUC & AUC & AUC \\ 
        & (Acc) & (F1 Score) & (mAP) & (Acc) & (F1 Score) & (mAP) \\ 
        \midrule
        DeepEns & 32.347 & 32.134 & 38.175 & 32.869 & 35.191 & 34.842 \\
        \textit{I}-EDL & 39.673 & 39.464 & 48.497 & 35.497 & 33.293 & 38.515 \\
        R-EDL & 38.137 & 39.423 & 48.869 & 39.275 & 33.896 & 35.110 \\
        GEM-FI & 39.001 & 38.943 & 47.684 & 37.144 & 35.080 & 37.534 \\
        ETN & 33.102 & 33.171 & 37.062 & 31.911 & 34.378 & 35.067 \\
        MaxLogit & 45.524 & 45.410 & 51.412 & 42.226 & 37.558 & 37.909 \\ \cmidrule(lr){1-7}
        \textbf{UEGMC-B} & \underline{65.441} & \underline{65.299} & \underline{71.751} & \underline{59.604} & \underline{57.764} & \underline{56.139} \\
        \textbf{UEGMC-P} & \textbf{75.017} & \textbf{75.035} & \textbf{82.712} & \textbf{72.154} & \textbf{68.101} & \textbf{69.759} \\
        \bottomrule
    \end{tabular}
    }
    \label{tab:tab3}
\end{table}

\begin{table}[!t]
    \centering
    \caption{Comparison of the morphology boundary uncertainty quantification across datasets with the ViT \cite{DBLP:conf/iclr/DosovitskiyB0WZ21} backbone. This table shows the relative Area Under Curve (AUC) (\%) $\uparrow$ when replacing a variable fraction of the most uncertain samples with the human supervised signal, while predictions of other samples remain unchanged. \textbf{Bold} and \underline{underlined} numbers denote the best and second best results respectively.}
    \resizebox{\columnwidth}{!}{
    \begin{tabular}{l|ccc|ccc}
        \toprule
        \multirow{3}{*}{Method} & \multicolumn{3}{c}{GalaxyMNIST} & \multicolumn{3}{c}{Galaxy10} \\ \cmidrule(lr){2-4} \cmidrule(lr){5-7}
        & AUC & AUC & AUC & AUC & AUC & AUC \\ 
        & (Acc) & (F1 Score) & (mAP) & (Acc) & (F1 Score) & (mAP) \\ 
        \midrule
        DeepEns & 37.514 & 37.893 & 40.953 & 37.318 & 45.221 & 44.104 \\
        \textit{I}-EDL & \underline{47.922} & 48.130 & \underline{58.351} & \underline{46.449} & \underline{45.865} & 46.501 \\
        R-EDL & 47.429 & \underline{49.914} & 57.025 & 40.983 & 42.117 & 45.990 \\
        GEM-FI & 46.411 & 46.401 & 56.454 & 45.055 & 45.447 & \underline{49.163} \\
        ETN & 36.376 & 36.408 & 40.426 & 37.458 & 45.195 & 44.617 \\
        \cmidrule(lr){1-7}
        \textbf{UEGMC-P} & \textbf{78.746} & \textbf{78.756} & \textbf{77.447} & \textbf{79.073} & \textbf{78.892} & \textbf{76.914} \\
        \bottomrule
    \end{tabular}
    }
    \label{tab:tab4}
\end{table}

\begin{figure}[!t]
    \centering
    \includegraphics[width=\columnwidth]{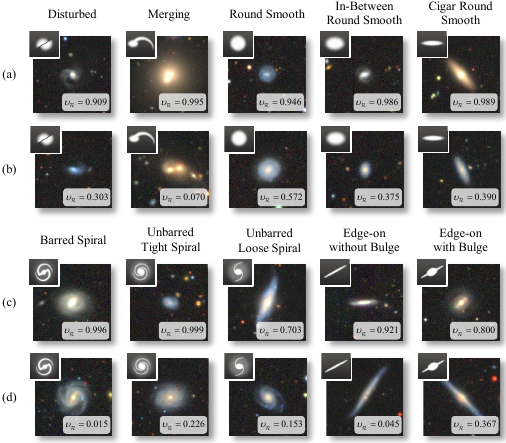}
    \caption{Examples with different morphology boundary uncertainty scores, quantified with UEGMC-P, where (a) and (c) represent samples with high uncertainty, while (b) and (d) represent samples with low uncertainty.}
    \label{fig:fig4}
\end{figure}

\subsection{Overall Performance}
For readability, we normalize the AUC between the best and worst possible results. The evaluation across four uncertainty types reveals that our UEGMC-P and its Bayesian baseline, UEGMC-B, consistently perform competitively compared with previous UQ methods on both datasets, as shown in Tables~\ref{tab:tab1}--\ref{tab:tab4}. Specifically, the results of model epistemic uncertainty in Table~\ref{tab:tab1} show that UEGMC-P efficiently identifies samples exceeding the cognitive capacity of the foundation model of a small size and achieves the best results. On data and modality aleatoric uncertainties in Tables~\ref{tab:tab2} and \ref{tab:tab3}, our UEGMC-P exhibits a robust capability to categorize ambiguities caused by instrumental degradation and multimodal misalignment, yielding substantial AUC improvements over previous baselines. 

Moreover, the results of the morphology boundary uncertainty in Table \ref{tab:tab4} demonstrate that UEGMC-P better captures the intrinsic evolutionary ambiguity of galaxies than previous methods. On average of the three metrics, our method outperforms the second-best method on datasets GalaxyMNIST and Galaxy10 by 26.254\% and 31.134\%, respectively. Overall, these results showcase that categorizing uncertainty into distinct, physically meaningful types improves the uncertainty evaluation of foundation models in GMC.

\begin{figure}[t]
    \centering
    \includegraphics[width=\columnwidth]{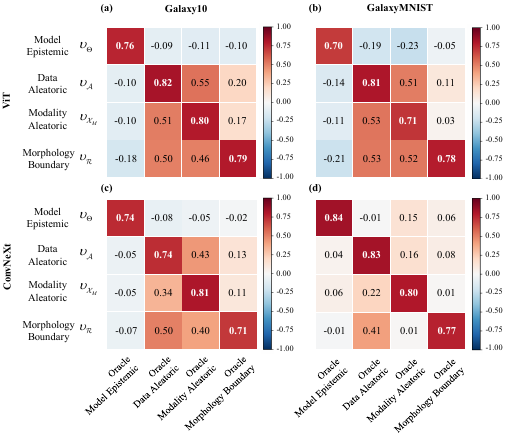}
    \caption{Pearson correlation between different uncertainty types of our UEGMC-P and Oracle. Red and blue colors represent positive and negative correlations, respectively, with color intensity corresponding to the magnitude of Pearson’s $r$, where scales are shown on the right.}
    \label{fig:fig5}
\end{figure}

\subsection{Qualitative Analysis and Ablation Experiments}
We show examples of galaxies with various levels of morphology boundary uncertainty in Figure~\ref{fig:fig4} to further assess the effectiveness of UEGMC-P. These examples differentiate between distinct and ambiguous galaxy morphologies. Specifically, examples assigned high uncertainty scores as shown in Figure \ref{fig:fig4} (a) and (c) exhibit visually ambiguous features, such as irregular structures and spiral arms. These characteristics represent evolutionary states that are typically misleading. In comparison, galaxies assigned low uncertainty scores, as shown in Figure \ref{fig:fig4} (b) and (d), display clearly defined structural boundaries. This contrast demonstrates that the morphology boundary uncertainty quantified by UEGMC-P moves beyond merely measuring model confidence to meaningfully capturing the intrinsic uncertainty in GMC.

UEGMC-P also demonstrates computational efficiency. As a post-hoc framework, it avoids the computational burden of retraining foundation models. The estimators are lightweight, requiring less than 6 million parameters and occupying a model size of under 23 MB. Consequently, it achieves an inference speed of over 0.5 million frames per second. These results demonstrate that UEGMC-P incurs only a marginal parameter overhead while delivering robust and highly competitive uncertainty evaluation in GMC.

We further ablate the extraction of feature representation layers. Given the streamlined architecture of our model, the primary configurable design choice is the depth of the extracted representations. We compare our strategy with three other extraction strategies, including relying solely on shallow layers, utilizing only deep layers, and including additional layers. When restricted to shallow layers, the model reduces to an AUC (Acc) of 69.460\% in terms of morphology
boundary uncertainty on the Galaxy10 dataset. Transitioning to deep layers alone decreases the evaluation AUC (Acc) to 71.223\%. In contrast, our proposed configuration yields the highest performance at 79.073\%, demonstrating its effectiveness. Notably, incorporating additional layers into our configuration provides no improvement, plateauing at 79.018\%.

\begin{figure}[t]
    \centering
    \includegraphics[width=\columnwidth]{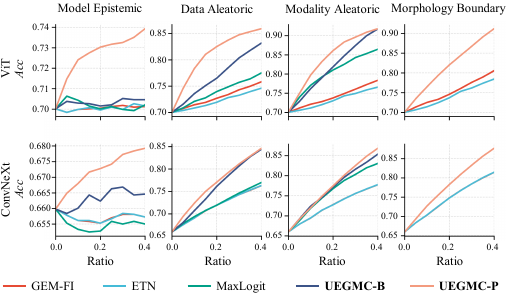}
    \caption{Comparison of accuracy performance gain while improving predictions selected with different UQ methods on the Galaxy10 dataset. The first row shows results that are evaluated with the ViT backbone setting. The second row shows results that are evaluated with the ConvNeXt backbone setting. Our UEGMC consistently shows promising performance compared with previous methods.}
    \label{fig:fig6}
\end{figure}

\subsection{Efficacy of Uncertainty Decomposition in GMC}
To assess whether our UEGMC-P accurately captures distinct types of uncertainty, we calculate the Pearson correlation between our estimated uncertainties and the Oracle results. As shown in Figure~\ref{fig:fig5}, the uncertainties quantified by UEGMC-P exhibit strong positive correlations with their corresponding Oracle measurements. This robust alignment indicates that our framework effectively identifies the target uncertainty types. This physically meaningful categorization provides a reliable basis for understanding model behavior in GMC.

To further demonstrate the efficacy of our uncertainty decomposition, we illustrate the performance improvements on the Galaxy10 dataset when the original estimates of a variable fraction of uncertain samples, as prioritized by different UQ methods, are substituted with improved ones in Figure~\ref{fig:fig6}. Across both the ViT and ConvNeXt backbones, sample refinement guided by UEGMC-P consistently yields the steepest and most substantial accuracy improvements compared to previous UQ methods. This performance gap demonstrates that UEGMC-P more precisely identifies the samples that are intrinsically vulnerable to misclassification, highlighting the effectiveness of uncertainty categorization for enhancing GMC.
\section{Conclusion}
In this paper, we address the problem of multi-source uncertainty quantification for GMC by proposing the UEGMC framework. We explicitly categorize the predictive uncertainty into four distinct types, including model epistemic, data aleatoric, modality aleatoric, and morphology boundary uncertainty, based on physical characteristics of galaxies and limitations in current GMC approaches. Our UEGMC framework enables foundation models to efficiently quantify uncertainties in a post-hoc manner through our lightweight and sampling-free UEGMC-P estimator. Extensive experiments on the Galaxy10 and GalaxyMNIST datasets with both ViT and ConvNeXt backbones demonstrate that UEGMC accurately identifies distinct sources of ambiguity, consistently showing competitive performance compared with previous methods. Crucially, UEGMC empowers astronomers to evaluate the origins of misclassification systematically and enables targeted interventions. Our UEGMC is currently evaluated on galaxy image data. Its effectiveness for broader and heterogeneous astrophysical data streams, such as time-series spectra, remains underexplored, though restricted by the current unavailability of suitably paired ground-truth data. We envision that extending our framework to incorporate such data could be an important next step. These extensions will further strengthen the reliability of foundation models, establishing comprehensive scientific credibility bounds for large-scale, AI-based astronomical discovery.

\bibliography{aaai2027}


\end{document}